\documentclass[conference]{IEEEtran}

\usepackage{amsmath,amsfonts}
\usepackage{algorithmic}
\usepackage{algorithm}
\usepackage{array}
\usepackage[caption=false,font=normalsize,labelfont=sf,textfont=sf]{subfig}
\usepackage{textcomp}
\usepackage{stfloats}
\usepackage{url}
\usepackage{verbatim}
\usepackage{graphicx}
\usepackage{cite}

\usepackage[utf8]{inputenc} 
\usepackage[T1]{fontenc}    
\usepackage{hyperref}       
\usepackage{booktabs}       
\usepackage{amsfonts}       
\usepackage{nicefrac}       
\usepackage{microtype}      
\usepackage{xcolor}         
\usepackage{multirow}
\usepackage{array}
\usepackage{xcolor}
\usepackage{amsmath}
\usepackage{geometry}
\usepackage{booktabs}
\usepackage{tabularx}
\usepackage{float}
\usepackage{wrapfig}
\usepackage{authblk}
\definecolor{mygray}{gray}{.7}
\IEEEoverridecommandlockouts
\usepackage{cite}
\usepackage{amsmath,amssymb,amsfonts}
\usepackage{algorithmic}
\usepackage{graphicx}
\usepackage{textcomp}
\usepackage{xcolor}
\def\BibTeX{{\rm B\kern-.05em{\sc i\kern-.025em b}\kern-.08em
    T\kern-.1667em\lower.7ex\hbox{E}\kern-.125emX}}
\begin{document}

\title{Advancing 3D Gaussian Splatting
Editing with Complementary and Consensus Information}

\author[1]{Xuanqi Zhang}
\author[2]{Jieun Lee}
\author[3]{Chris Joslin}
\author[1]{Wonsook Lee}
\affil[1]{University of Ottawa}
\affil[2]{Hansung University}
\affil[3]{Carleton University}

\maketitle

\begin{abstract}
We present a novel framework for enhancing the visual fidelity and consistency of text-guided 3D Gaussian Splatting (3DGS) editing.
Existing editing approaches face two critical challenges: inconsistent geometric reconstructions across multiple viewpoints, particularly in challenging camera positions, and ineffective utilization of depth information during image manipulation, resulting in over-texture artifacts and degraded object boundaries.
To address these limitations, we introduce:
1) A complementary information mutual learning network that enhances depth map estimation from 3DGS, enabling precise depth-conditioned 3D editing while preserving geometric structures.
2) A wavelet consensus attention mechanism that effectively aligns latent codes during the diffusion denoising process, ensuring multi-view consistency in the edited results.
Through extensive experimentation, our method demonstrates superior performance in rendering quality and view consistency compared to state-of-the-art approaches. The results validate our framework as an effective solution for text-guided editing of 3D scenes.

\end{abstract}

\begin{IEEEkeywords}
3D editing, 3D Gaussian splatting, depth enhancement, diffusion models.
\end{IEEEkeywords}

\section{Introduction}
3D editing has emerged as a promising technique in applications such as augmented reality, virtual reality, and visual effects. 
Existing approaches for 3D scene editing can be primarily divided into two categories: direct 3D editing methods and 2D-to-3D lifting methods \cite{lu2024advances,luo2024trame,chen2024dge}.
Direct 3D editing typically operates on 3D data representations such as meshes or point clouds. 
However, these approaches face significant challenges, including the capture and annotation of 3D datasets and difficulties in achieving high-fidelity results after 3D editing.

To address these limitations, researchers have explored a two-phase paradigm: editing multi-view 2D images first, followed by lifting these edited 2D views into 3D. 
This approach often mitigates the limitations of direct 3D editing while improving visual quality. 
During the image editing phase, text-guided diffusion models have gained significant attention for their intuitive interface, allowing users to modify images through simple text instructions.
Techniques such as ControlNet \cite{zhang2023adding} and T2I-Adapter \cite{mou2024t2i} enhance the controllability of these models by incorporating various conditioning inputs like segmentation maps and depth maps.

In the subsequent 3D reconstruction phase, recent methods like InstructN2N \cite{haque2023instruct} edit images and update 3D datasets based on Neural Radiance Fields (NeRF). However, NeRF's training and rendering processes are often time-consuming, posing challenges for real-time applications.
The 3D Gaussian Splatting (3DGS)  {\cite{kerbl20233d}} is an emerging technique which offers an efficient alternative by employing  {explicit neural representations} for 3D reconstruction.

 Ensuring consistency across edited multi-view images remains a critical challenge for 2D-to-3D lifting approaches. Several consistency-aware methods have been developed to address this issue \cite{wu2024gaussctrl,dong2024vica}. For example, GaussCtrl\cite{wu2024gaussctrl} utilizes ControlNet for depth-conditioned editing across all images.
However, the depth maps generated by 3DGS frequently contain spurious texture patterns that represent surface details rather than actual geometric structures. These erroneous patterns, when used directly, can introduce undesirable artifacts in downstream tasks \cite{chen2024intrinsic}.

Therefore, enhancing depth map quality is essential for achieving superior image editing results in 3D scenes. 
Inspired by the previous cross-modal enhancement research \cite{dong2022learning}, we develop the Complementary Information Mutual Learning Network (CIMLN). 
This network leverages complementary information from multi-view images to refine depth maps, enabling precise depth information which is used as an additional condition for diffusion-based image editing in ControlNet. 
We further propose a {Wavelet Consensus Attention (WCA) module} to ensure advanced multi-view consistency and extract consensus information from both the spatial and frequency domains.
The WCA module enhances multi-view consistency by replacing ControlNet's self-attention with wavelet-based attention, enabling latent code alignment during the diffusion denoising process. 
Therefore, we advance complementary and consensus information \cite{zhang2023multi, shen2024dual} for 3D Gaussian editing.
This architectural modification facilitates seamless coordination across viewpoints, resulting in more coherent multi-view editing results.

Our main contributions are summarized as follows:

\begin{itemize} 
\item We propose an efficient 3D editing method based on 3DGS that incorporates complementary and consensus information learning for improved performance.

\item We develop a self-supervised complementary information learning network that extracts complementary details from 3D depth information and multi-view images to refine depth maps.

\item We design a wavelet consensus attention for latent code alignment module that extracts consensus information from both spatial and frequency domains and ensures consistent editing results.

\item Extensive experiments demonstrate that our proposed framework achieves superior performance compared to state-of-the-art methods.
\end{itemize}

\begin{figure*}
	\centering
        \includegraphics[width=1\linewidth]{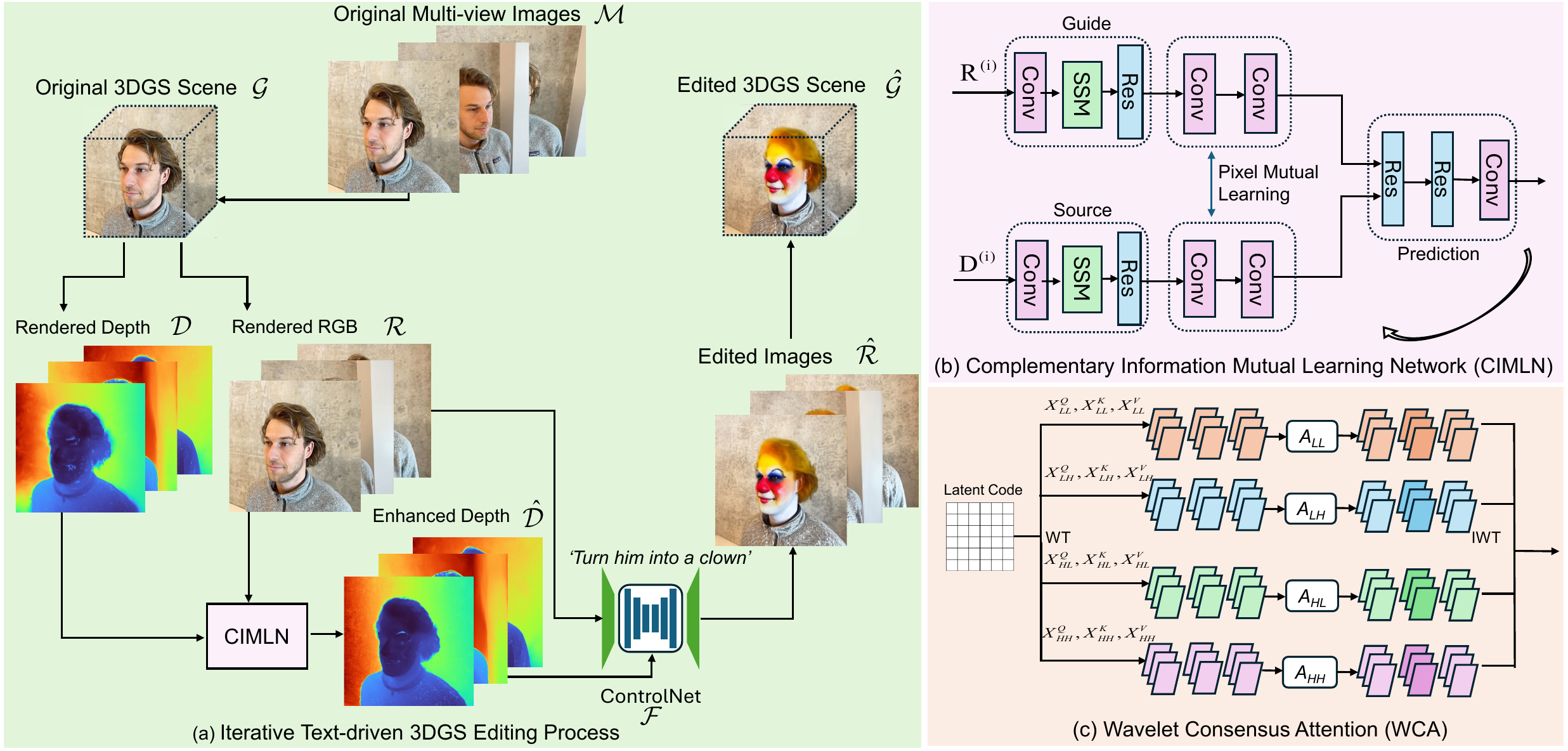}
	\caption{Overview of the proposed 3D Gaussian splatteing editing system. (a) Multi-view images $\mathcal{M}$ are used to train a 3DGS model, from which the rendered depth and RGB images can be obtained from a certain viewpoint. { (b) Rendered depth maps and RGB images are processed through Source and Guide branches respectively. Through pixel mutual learning and downsampling, the system enables self-supervised training. (c) The framework replaces ControlNet's image self-attention with WCA to better align images with reference views.}}
	\label{flowchart} 
\end{figure*}

\section{Related Work}
\textbf{3DGS Editing}
3DGS has achieved state-of-the-art results in neural novel view synthesis; however, the development of effective editing methods for this representation remains a fundamental challenge in the field \cite{chen2024dge,wu2024gaussctrl}. Instruct-GS2GS \cite{igs2gs} attempt to employ 3DGS to improve editing efficiency while maintaining consistency across views. GaussianEditor \cite{chen2024gaussianeditor} introduces a mesh-based neural field approach, which allows precise localization and separation of texture and geometry for flexible editing.

\textbf{Depth Enhancement}
Techniques for enhancing depth maps can be broadly classified into optimization-based and learning-based approaches\cite{zhong2023guided}. Optimization-based methods often rely on formulating depth enhancement as an ill-posed problem, incorporating priors to make it solvable\cite{wang2023rgb,de2022learning}. 
Learning-based approaches leverage neural networks to refine depth maps from color-depth pairs \cite{chen2024intrinsic}. For example, SGNet \cite{wang2024sgnet} employs a structure-guided network with gradient-frequency awareness to enhance depth details, achieving superior performance with sharper boundaries and better detail preservation.

\section{Method}
Our proposed method utilizes 3DGS for efficient 3D editing by addressing two core challenges: multi-view consistency and depth map refinement. We achieve this through two key components: (1) a Complementary Information Mutual Learning Network (CIMLN) for precise depth enhancement, and (2) a Wavelet Consensus Attention (WCA) mechanism for latent code alignment in ControlNet.

\subsection{Reconstruction-Based 2D-to-3D Lifting Using 3DGS}
We adopt the 3DGS representation for efficient 3D scene reconstruction, modeled as a mixture of Gaussians:
\begin{equation} \label{gs}
\begin{array}{lr}
\mathcal{G} = \{(\sigma_i, \mu_i, \Sigma_i, c_i)\}_{i=1}^{ {N}},
\end{array}
\end{equation}
where $\sigma_i \geq 0$ denotes the opacity, $\mu_i \in \mathbb{R}^3$ is the mean, $\Sigma_i \in \mathbb{R}^{3 \times 3}$ is the covariance matrix, $c_i$ represents the color of each Gaussian and $N$ is the number of Gaussian ellipsoids.
Although 3DGS provides a computationally efficient representation, directly modifying these parameters can produce noisy results due to entangled geometry and texture attributes \cite{sella2023vox}.

\begin{figure*}
	\centering
        \includegraphics[width=1\linewidth]{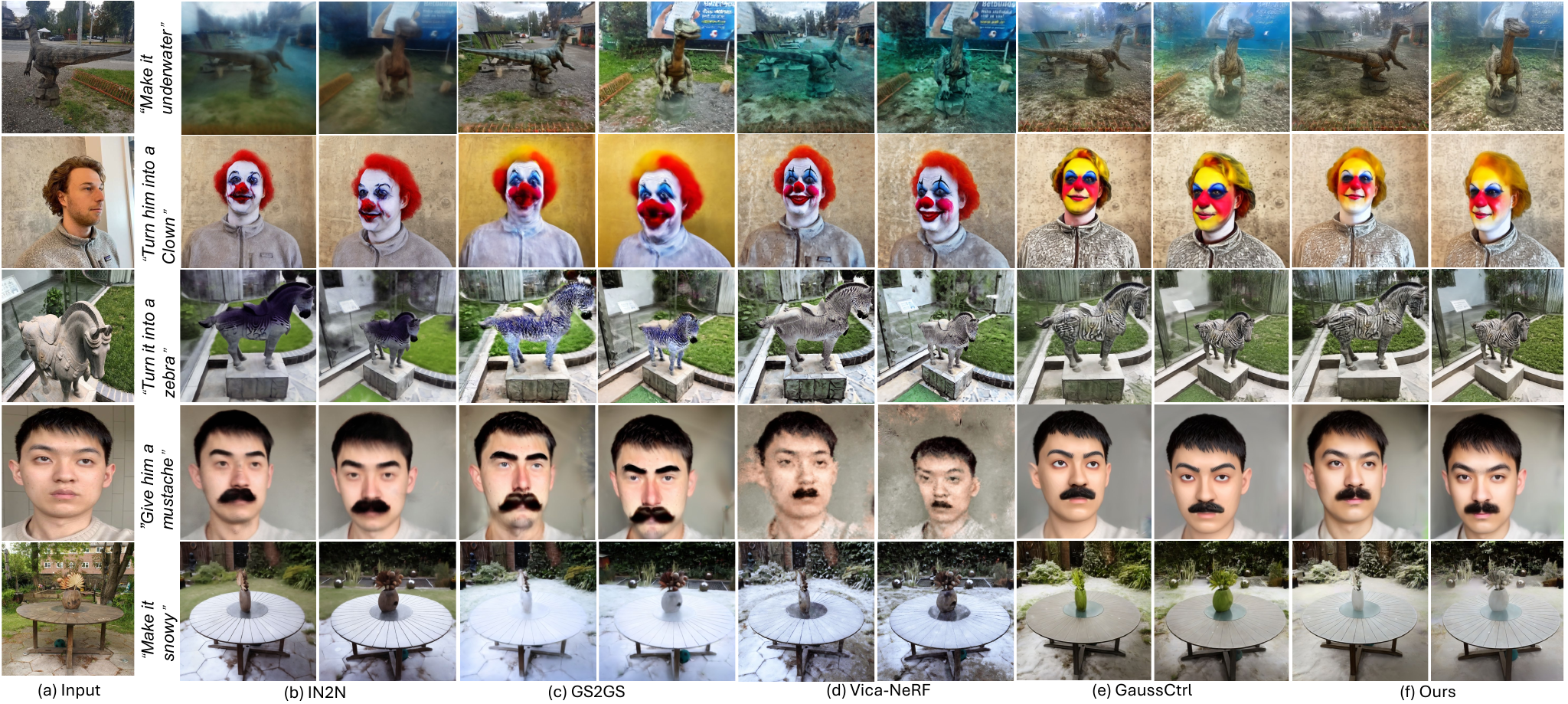}
	\caption{Visual comparison of text-drive 3D editing on different scenes. The inconsistency and noises are decreased in detail.}
	\label{Overall} 
\end{figure*}

As illustrated in Fig. \ref{flowchart}, we render RGB images $\mathcal{R}$ and their corresponding depth maps $\mathcal{D}$ from the 3DGS model. 
These outputs are used as input to ControlNet for depth-conditioned image editing. 
However, depth maps generated by 3D Gaussian often lack sufficient precision, leading to inconsistent editing. To address this, we propose CIMLN for depth refinement, ensuring more accurate and reliable depth maps.

 After the enhanced depth maps $\mathcal{D}$ are obtained, we send them into the ControlNet $\mathcal{F}$ along with the rendered RGB images and a text instruction $p$.
During DDIM inversion, the latent code $z_0$ is inverted to Gaussian noise $z_t$. The predicted noise is:
\begin{equation}
{e}^t = \mathcal{F}(z_t; t, p, \mathcal{D})
\end{equation}
\begin{align}
z_{t+1} = \sqrt{\alpha_{t+1}} z_t - \frac{\sqrt{1 - \alpha_t} \cdot {e}^t}{\sqrt{\alpha_t}} + \sqrt{1 - \alpha_{t+1}} {e}^t,
\end{align}
where $\alpha_t$ denotes the opacity. 
To align the weight between different latent codes,
WCA is employed to align them from both spatial and frequency domains and generates new $\hat{e}^t$. The denoise process is:
\begin{align}
{z}_{t-1} &= \sqrt{\alpha_{t-1}} {z}_t - \frac{\sqrt{1 - \alpha_t} \cdot \hat{e}^t}{\sqrt{\alpha_t}} + \sqrt{1 - \alpha_{t-1}} \hat{e}^t.
\end{align}

After we obtain ${z}_{0}$, it will be decoded to edited image $\hat{R}$ by the VAE decoder. In the end, the edited images will be used to train the edited 3DGS model.

\subsection{Complementary Information Mutual Learning Network}

To estimate depth from 3DGS, 3D Gaussians are projected into camera space as 2D Gaussians.
The 2D Gaussians are globally sorted by depth per pixel, and the depth estimates $\mathbf{\hat{D}}$ through the discrete volume rendering approximation:
\begin{equation}
\mathbf{\hat{D}} = \sum_{i \in \mathbf{N}} d_i \alpha_i \prod_{j=1}^{i-1} (1 - \alpha_j),
\end{equation}
where $d_i$ denotes the depth of the $i$-th Gaussian in view space, and $\alpha_t$ denotes its opacity.
While this method approximates per-pixel depth, its accuracy is limited in regions with overlapping Gaussians.
Additionally, the detailed textures rendered from Gaussians are undesirable noise for the depth map.

To enhance the depth maps rendered by the 3DGS model, we propose CIMLN which is designed to extract edge information from the depth image, while suppressing texture details from the color image. By learning complementary information from the rendered images, CIMLN produces enhanced depth maps with sharper edges and higher spatial resolution.

As shown in Fig. \ref{flowchart} (b), rendered depth maps $\mathcal{D}$ and rendered RGB images $\mathcal{R}$ are sent as the inputs to the source branch and guide branch respectively.
Given input $\mathbf{I}_s \in \mathbb{R}^{H \times W \times 1}$ and $\mathbf{I}_g \in \mathbb{R}^{H \times W \times 3}$, 
CIMLN uses a simplified state space model \cite{gu2023mamba,zhang2024llemamba} (SSM) to extract features in the branches with the global perceptive field, as shown in the following equation.
\begin{equation} \label{ssm}
\begin{array}{lr} 
\mathbf{F_t} = \text{Conv2d}(\mathbf{F}), \\
\mathbf{X} = \text{SILU}(\text{Conv1d}(\mathbf{F_t})), \quad \mathbf{Y} = \text{SILU}(\text{Conv1d}(\mathbf{F_t})), \\
\mathbf{X}_t = \text{LN}(\text{SSM}(\mathbf{X})), \quad \mathbf{X}_{\text{out}} = \text{Linear}(\mathbf{X_t} \odot \mathbf{Y}),
\end{array}
\end{equation}

where LN represents LayerNorm and SiLU is an activation function. 
Compared to traditional CNN, SSM efficiently captures global information through the ability to capture long-range dependencies.

\begin{figure*}
	\centering
        \includegraphics[width=1\linewidth]{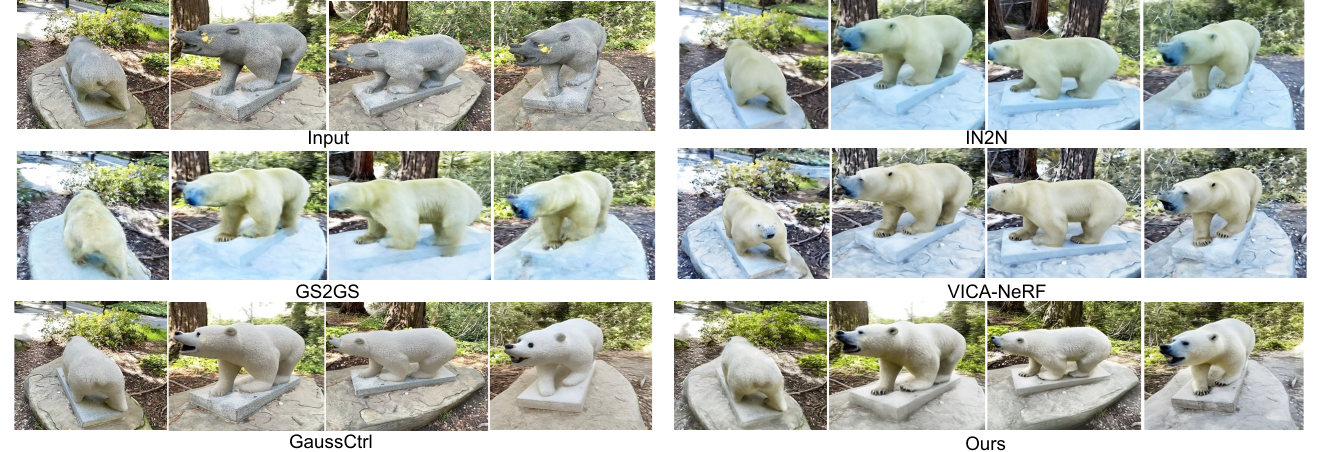}
	\caption{Visual comparison of prompt 'Turn it into a polar bear'. Our method maintains multi-view consistency for rendered images.}
    \label{bear}
\end{figure*}

\begin{figure}	
    \centering        
  \includegraphics[width=0.8\linewidth]{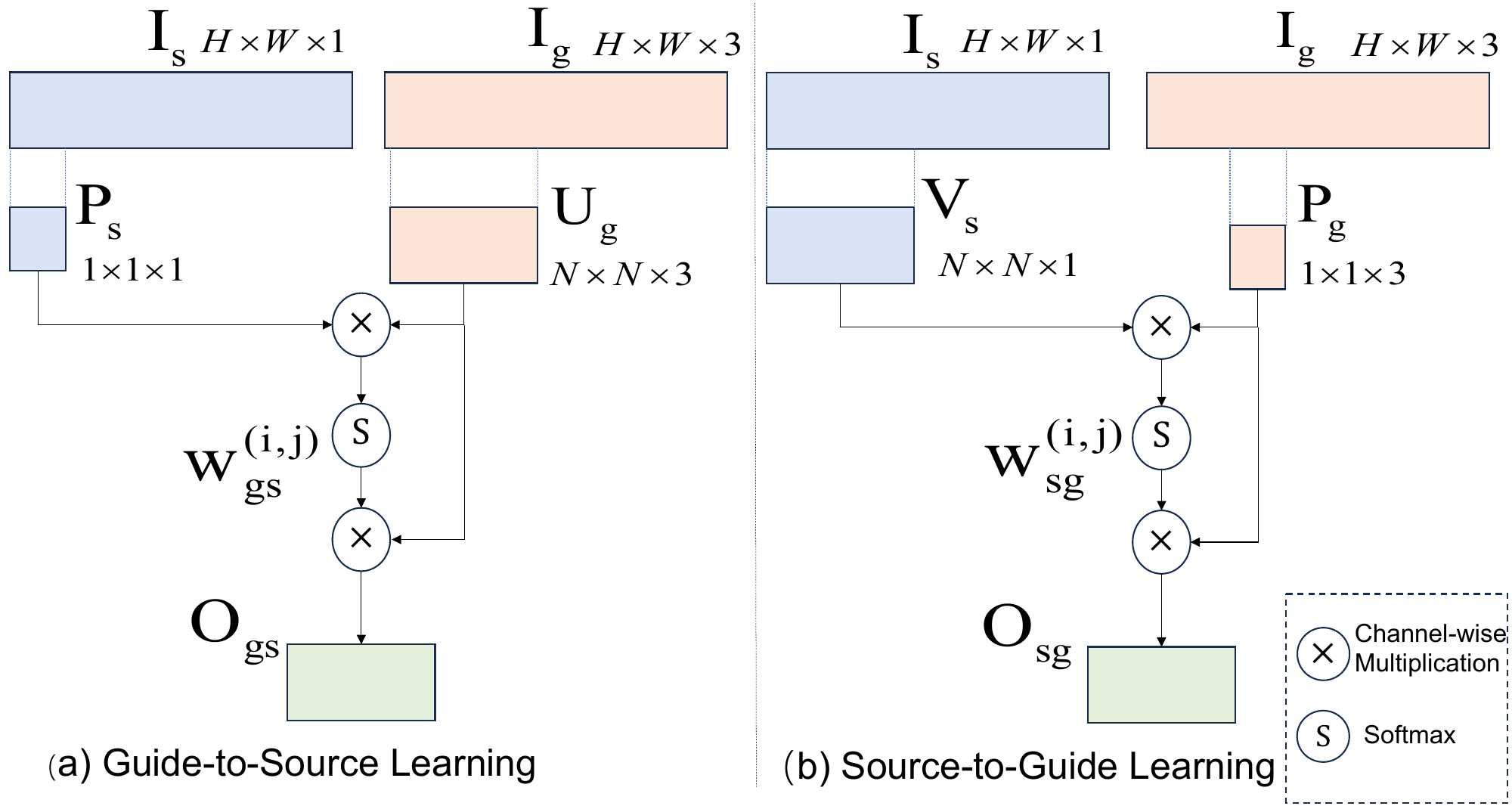}	
    \caption{Detailed design of Pixel Mutual Learning.} 	
    \label{CIMLN}
\end{figure}

As shown in Fig. \ref{CIMLN}, the guide-to-source and source-to-guide branches exchange features through a Pixel Mutual Learning module, which aligns pixel-level information between RGB and depth.
In Fig. \ref{CIMLN}~(a), we first extract a  {depth} $\mathbf{P}_s$ from source image and  {an RGB }$\mathbf{U}_g$ from guide image, and then perform matrix multiplication for them.
A weight filter is employed to evaluate the correlation value between  {a pixel in source} $p_s$ and each pixel in  { guide }$\mathbf{U}_{g(i,j)}$:
\begin{equation} \label{weight}
\begin{array}{lr}
\mathbf{w}_{gs}^{(i,j)} = \text{Softmax}\left( \left(\mathbf{U}_{g_{(i,j)}}\right)^\top p_{s} \right), \\
\mathbf{O}_{gs}  = \mathbf{U}_{g_{(i,j)}}, \mathbf{w}_{gs}^{(i,j)}.
\end{array}
\end{equation}
The source-to-guide pixel mutual learning follows the same process. 
Lastly, the convolution layers and residual blocks are introduced for the final prediction to obtain the output image.
Output Image is downsampled as $\mathbf{I}_{out}$ for self-supervised training.
The training loss is:
\begin{equation} \label{loss}
\begin{array}{lr}  \mathcal{L}_{total}= \lambda\mathcal{L}_{L1}+\gamma\mathcal{L}_{ba}
=\lambda \| \mathbf{I}_{out} - \mathbf{I} \|^2_F + \\ \gamma\left( \nabla_x \mathbf{I}^h - \nabla_x \mathbf{I}^{out}_{m-1} \right) \odot \left( \nabla_y \mathbf{I}^h - \nabla_y \mathbf{I}^{out}_{m-1} \right)
,
\end{array}
\end{equation}
where  $\mathcal{L}_{L1}$ measures the pixel-wise errors between the output image and the ground-truth image.
$\mathcal{L}_{ba}$ represents boundary-aware loss
to generate sharper boundaries.
This approach simultaneously achieves enhanced output and maintains modality fidelity through a fully self-supervised learning framework.

\subsection{Wavelet Consensus Attention-Based Latent Code Alignment}

To ensure multi-view consistency during text-driven
diffusion editing, we introduce a WCA mechanism to align the latent codes within
the ControlNet. This mechanism increases the size of the
layer’s receptive field and improves its ability to capture
low-frequency features \cite{finder2025Wavelet}. As shown in Fig. \ref{flowchart} (c), the
wavelet decomposition is employed on latent codes and
their corresponding attention:
\begin{equation} \label{attention}
\begin{array}{lr}
\mathbf{A}_{C}(\mathbf{X}_{C}^{Q}, \mathbf{X}_{C}^{K}, \mathbf{X}_{C}^{V}) = \text{Softmax}(\frac{{\mathbf{X}_{C}^{Q}}{\mathbf{X}_{C}^{K}}^T}{\sqrt{\alpha}}) \mathbf{X}_{C}^{V},
\end{array}
\end{equation}
where $C \in {LL, LH, HL, HH}$ represents wavelet component of wavelet transformation (WT). $\mathbf{X}^{Q} ,\mathbf{X}^{K},\mathbf{X}^{V}$ are the linear projections to obtain query, key, and value. Inverse wavelet transformation (IWT) is performed to obtain the attention of two latent codes.
We incorporate the self-attention methodology to maintain distinctive features:
\begin{equation} \label{align_attention}
\begin{array}{lr}
\mathbf{Attn}_{i} = \lambda \cdot \mathbf{A}_{i,i} + (1 - \lambda) \cdot \frac{1}{N_r} \sum_{ {j=1}}^{N_r} \mathbf{A}_{i,j},
\end{array}
\end{equation}
where $\mathbf{A}_{i,i}$ represents self-attention, $\mathbf{A}_{i,j}$ denotes WCA, and $\lambda \in [0, 1]$. This mechanism aligns the appearance of all edited images to the reference views via latent codes across both spatial and frequency domains, contributing to appearance consistency and artifact removal.

\begin{table}[h]
\centering
\caption{Quantitative Comparison of Reconstruction-Based Methods (Best Results in Bold).
}
\begin{tabular}{llccc}
\toprule
\multirow{1}{*}{Methods}
& PSNR  $\uparrow$ & RMSE $\downarrow$ & LPIPS $\downarrow$ & Time(min) $\downarrow$ \\
\midrule
IN2N & 42.35 & 0.492 & 0.319 & $\sim${82} \\
IGS2GS & 40.68 & 0.559 & 0.402 & $\sim${29} \\
GaussCtrl & 50.53 & 0.121 & 0.205 & \textbf{$\sim${15}} \\
ViCA-NeRF & 48.21 & 0.215 & 0.182 & $\sim${61} \\
Ours & \textbf{51.82} &\textbf{0.118} & \textbf{0.145} & {$\sim${28}} \\
\bottomrule
\end{tabular}
\label{psnr}
\end{table}

\begin{table}[h]
\centering
\caption{CLIP$_{dir}$ Comparison of Text-to-Image Consistency (Best Results in Bold).}
\resizebox{0.48\textwidth}{!}{
    \begin{tabular}{lccccc}
        \toprule
        \multirow{2}{*}{Scene} & \multicolumn{5}{c}{CLIP Text-Image Directional Similarity $\uparrow$ } \\ \cmidrule(r){2-6} 
        & IN2N & IGS2GS & GaussCtrl &ViCA-NeRF  & Ours \\ \midrule
        Bear Statue & 0.1019 & 0.1165 & 0.1388 & 0.1104 & \textbf{0.1428} \\
        Dinosaur    & 0.1466 & 0.1490 & 0.1584 & 0.0723 & \textbf{0.1654}   \\
        Garden      & \textbf{0.3027} & 0.1663 & 0.2891 & 0.2903& 0.2911 \\
        Stone Horse & 0.1654 & 0.1947 & 0.2268 & 0.1926& \textbf{0.2317} \\
        Fangzhou    & 0.1598 & 0.2032 & 0.1887 & 0.1809& \textbf{0.2153} \\
        Face        & 0.1332 & 0.1357 & 0.1503 & 0.1119& \textbf{0.1648} \\     \bottomrule
    \end{tabular}
    }    
    \label{table:clip}
\end{table}

\section{Experiment}
\subsection{Experiment Setup}
Following IN2N \cite{haque2023instruct}, we conducted experiments from several established datasets, including
Mip-NeRF~\cite{barron2022mip},
BlendedMVS~\cite{yao2020blendedmvs}, and
NeRF-Art~\cite{wang2023nerf}.
To process the datasets, we adopted the camera path extraction method suggested by NeRFStudio \cite{tancik2023nerfstudio}.
All the experiments are conducted on the NVIDIA GeForce RTX 4090 GPU with PyTorch framework.

Our evaluation metrics include:
CLIPdir (CLIP Text-Image Directional Similarity) quantifies the semantic consistency between text instructions and visual changes in the edited images.
PSNR (Peak Signal-to-Noise Ratio) measures the fidelity of the reconstructed views.
RMSE (Root Mean Square Error) assesses reconstruction accuracy.
LPIPS (Learned Perceptual Image Patch Similarity) evaluates the perceptual similarity between 2D edited images and the rendered 3D outputs, emphasizing human visual perception.

\subsection{Qualitative Results}
We conducted extensive qualitative evaluations to compare the visual editing effects of our method against state-of-the-art techniques, including IN2N \cite{haque2023instruct}, GS2GS \cite{igs2gs}, GaussCtrl \cite{wu2024gaussctrl}, and ViCA-NeRF \cite{dong2024vica}.
As shown in Fig. \ref{bear}, the proposed method consistently produces high-quality images of the bear with superior multi-view consistency.
Other methods, such as GS2GS and VICA-NeRF, struggle to produce multi-view consistent results after editing.

 For example, when applying the text prompt "Give him a mustache" to the Fangzhou scene, as shown in Fig. \ref{Overall}, IN2N \cite{haque2023instruct} and ViCA-NeRF \cite{dong2024vica} generated noisy and inconsistent outputs.
GaussCtrl yielded better results but introduced significant style changes, potentially conflicting with user expectations.
The supplementary materials contain an extensive collection of alternative viewpoints and angles of these 3D scenes.
Our method effectively preserved the intended transformation while maintaining background consistency.

\begin{figure}
    \centering
    \includegraphics[width=1\linewidth]{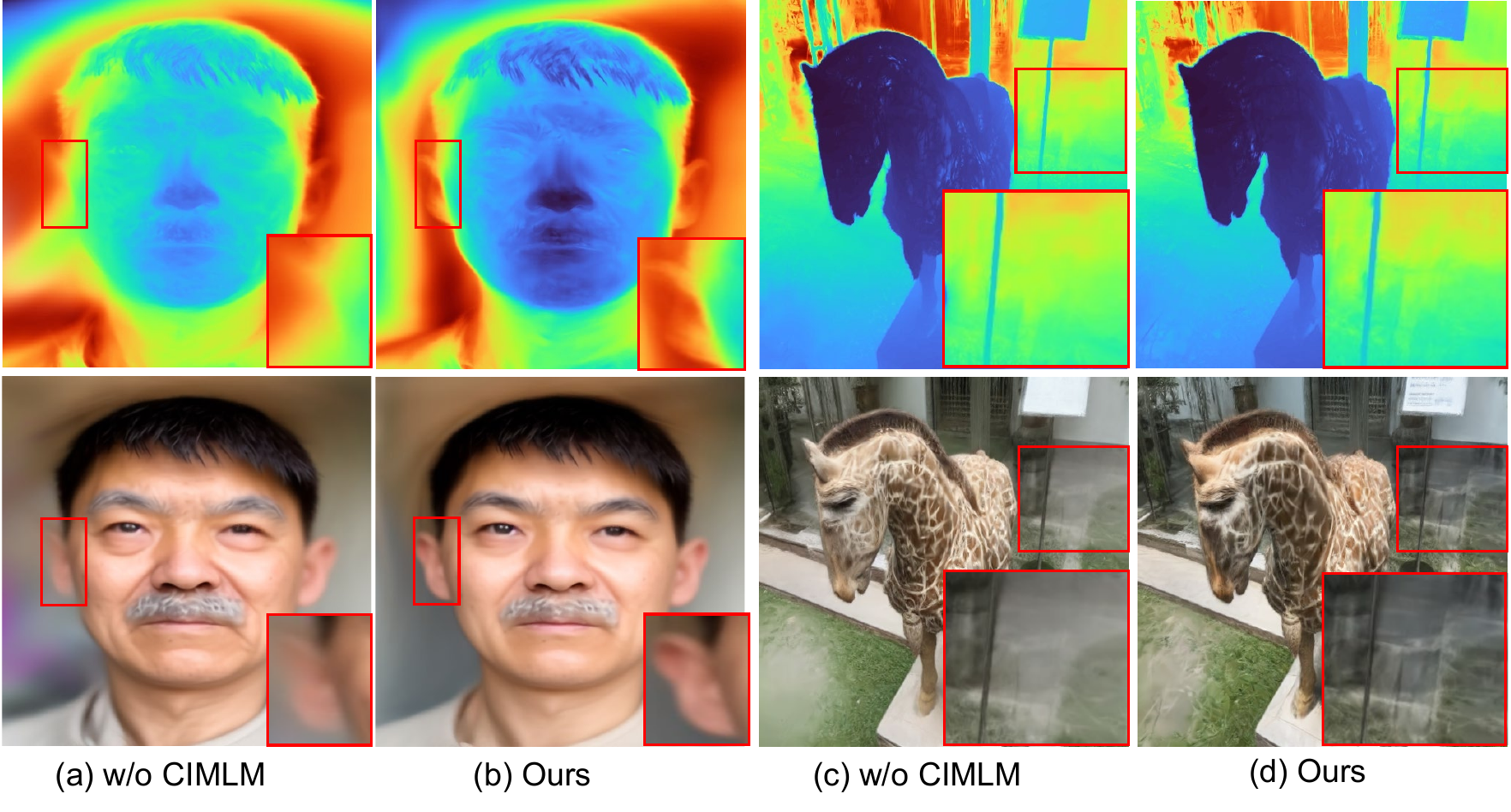}
    \caption{Ablation study on the effect of the proposed CIMLN module.}
    \label{fig:ablation1}
\end{figure}

\begin{figure}
    \centering
    \includegraphics[width=1\linewidth]{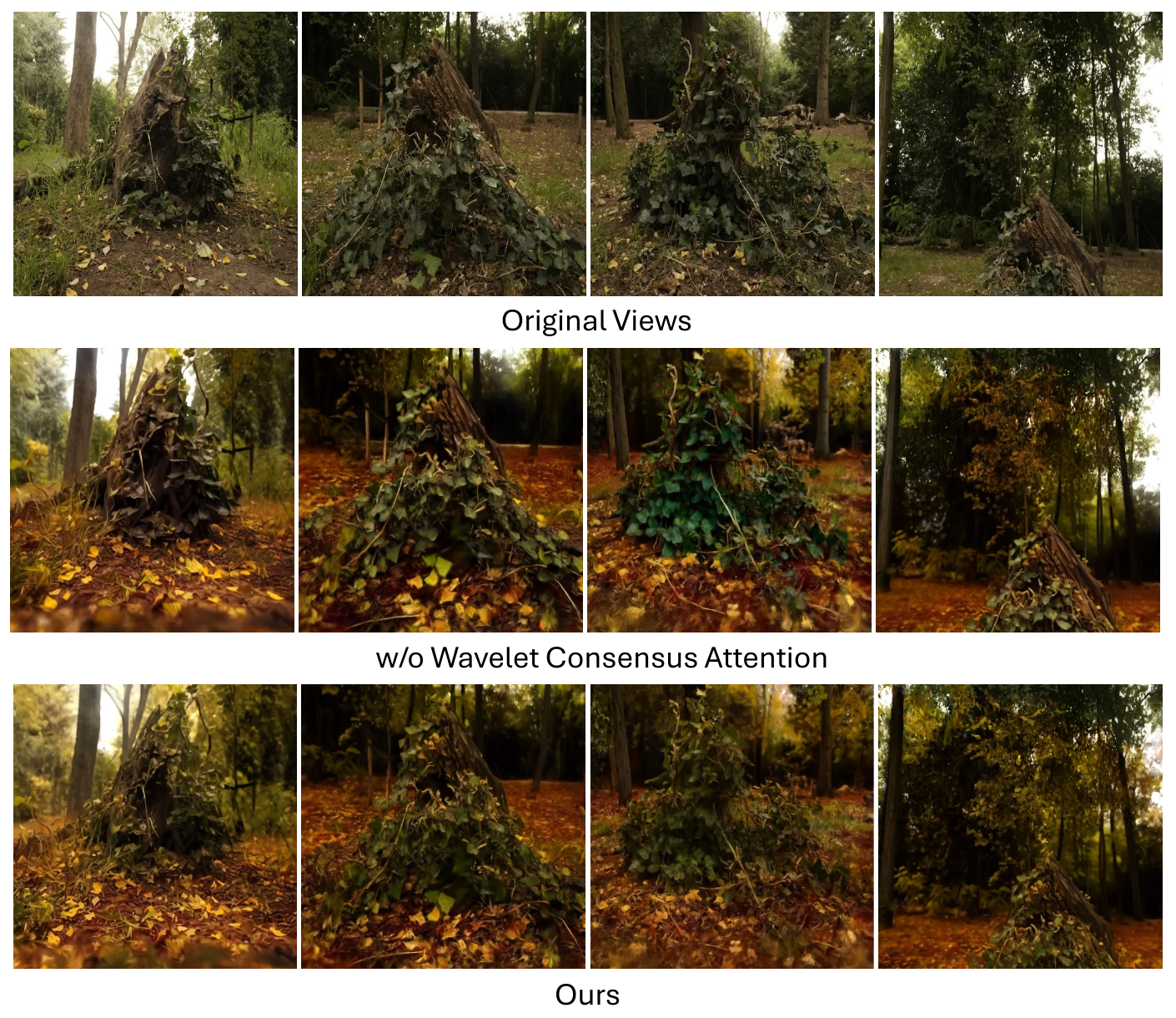}
    \caption{Ablation study on the effect of the proposed WCA module.}
    \label{fig:ablation2}
\end{figure}

\subsection{Quantitative Results}
The quantitative performance of different methods is summarized in Table~\ref{psnr} and Table~\ref{table:clip}.
Table~\ref{psnr} summarizes the average metrics across six scenes, which shows that our method outperformed competing approaches in PSNR, RMSE, and LPIPS across these datasets. Specifically, our method demonstrates superior performance, achieving a 1.29 dB improvement in PSNR compared to GaussCtrl.  By leveraging complementary information from RGB and depth images, our method accurately learns depth information for diffusion generation, resulting in high-quality edited images.
Table \ref{table:clip} reports CLIPdir scores, where our method achieved the highest semantic alignment between text prompts and image transformations across most datasets.

Additionally, we evaluated the overall editing time to assess the suitability of the models for real-time applications. As shown in Table~\ref{psnr}, NeRF-based methods (e.g., IN2N, ViCA-NeRF) were significantly slower, whereas our method provided a balance between speed and performance due to the efficiency of the 3D Gaussian Splatting representation. 
One-time dataset editing also contributed to faster convergence compared to iterative update methods (e.g., IN2N, GS2GS, and VICA-NeRF). 

\section{Ablation Study} 
\textbf{Effect of CIMLN} To evaluate the impact of our proposed CIMLN, we conducted ablation studies on the Face scene with the text prompt "Make him older" and the Stone Horse scene with the text prompt "Turn it into a giraffe".
As shown in Fig. 5, the inclusion of CIMLN results in depth maps with sharp edges and reduced noise in rendered RGB images.
The results indicate that the enhanced depth map by CIMLN can result in better editing results.

\textbf{Effect of WCA}
As shown in Fig. 6, when editing the Stump scene with the text prompt "Make it in autumn," the absence of the WCA module causes the leaf colors to vary significantly across different viewpoints, which is undesirable for multi-view consistency in 3D scenes. In contrast, our method, which incorporates WCA, maintains multi-view consistency and achieves high color fidelity, ensuring a cohesive autumn background in the forest across all views.

\section{Conclusion}  
This paper proposed a novel framework for 3D scene editing based on 3DGS. Our approach addresses two key challenges of depth map refinement and multi-view consistency. The CIMLN enhances depth maps by leveraging complementary information from multi-view RGB and depth inputs, while the WCA mechanism ensures consistent edits across views by aligning latent codes in both spatial and frequency domains. The proposed framework achieves high-quality visual results, reliable multi-view consistency, and efficient editing performance.

\bibliographystyle{IEEEbib}
\bibliography{icme2025references}

\end{document}